\documentclass{article}
\usepackage[utf8]{inputenc}
\usepackage[a4paper, margin=1in]{geometry}
\usepackage{graphicx}
\usepackage[numbers]{natbib}
\usepackage[colorlinks=true, allcolors=blue,hypertexnames=false,pagebackref=true]{hyperref}
\usepackage[font=small,labelfont=bf]{caption}

\usepackage{amsmath}
\usepackage{amsfonts}
\usepackage{authblk}

\usepackage{booktabs}
\usepackage[T1]{fontenc}

\usepackage{svg}
\usepackage{dsfont} %for mathds{1}
\usepackage{multirow} %for table
\usepackage{array}  %for table
\usepackage{arydshln} %for table
\usepackage{pbox} % line breaks in tables
\usepackage{booktabs,tabularx}
\usepackage[table]{xcolor} % for gray header
\usepackage{siunitx}
\usepackage{fontawesome5}
\newcommand{\catP}{\faUser}    % �� patient
\newcommand{\catA}{\faCamera}  % �� acquisition
\newcommand{\catM}{\faCog}     % ⚙️ model

\newcommand{\noP}{\phantom{\faUser}}
\newcommand{\noA}{\phantom{\faCamera}}

\newcommand{\numpm}[2]{%
  \num{#1}\,$\pm$\,\num{#2}%
}
\newcolumntype{N}{>{\centering\arraybackslash}p{1.95cm}}
\newcolumntype{Y}{>{\raggedright\arraybackslash}X}

\usepackage{changepage}                 % adjust margins for selected portions
\usepackage{placeins}
\let\Oldsection\section
\renewcommand{\section}{\FloatBarrier\Oldsection}
\let\Oldsubsection\subsection
\renewcommand{\subsection}{\FloatBarrier\Oldsubsection}
\let\Oldsubsubsection\subsubsection
\renewcommand{\subsubsection}{\FloatBarrier\Oldsubsubsection}
\newlength{\offsetpage}
{\end{adjustwidth}}
\usepackage{tocloft} % for fiddling with the ToC, LoF and LoT
\cftpagenumbersoff{figure} % no page numbers
\makeatletter
\renewcommand{\fnum@figure}[1]{\figurename\ \thefigure} % remove the : after number
\makeatother

\usepackage{marvosym} % for email symbol \Letter

\usepackage{setspace}
\newcommand*{\MinNumber}{40}%
\newcommand*{\MaxNumber}{200}%

\newcommand{\ApplyGradient}[1]{%
  \pgfmathsetmacro{\PercentColor}{100*(#1-\MinNumber)/(\MaxNumber-\MinNumber)}%
  \edef\x{\noexpand\cellcolor{black!\PercentColor}}\x\textcolor{black}{#1}%
}
\newcolumntype{R}{>{\collectcell\ApplyGradient}{c}<{\endcollectcell}}

\usepackage{xcolor}
\newcommand{\new}[1]{\textcolor{black}{#1}}

\title{Detection and Validation of Hidden Subgroup Disparities in Clinical AI}

\author[1,2,3,\Letter]{Alceu Bissoto}
\author[1,2,3]{Trung-Dung Hoang}
\author[1,2,3]{Tim Flühmann}
\author[4]{Nina Weng}
\author[5]{Susu Sun}
\author[5,6]{Christian F. Baumgartner}
\author[1,2,3,\Letter]{Lisa M. Koch}

\affil[1]{Department of Digital Medicine, University of Bern, Switzerland}
\affil[2]{Department of Diabetes, Endocrinology, Nutritional Medicine and Metabolism, Bern University Hospital, Switzerland}
\affil[3]{Diabetes Center Berne, Switzerland}
\affil[4]{Technical University of Denmark, Denmark}
\affil[5]{Cluster of Excellence: Machine Learning - New Perspectives for Science, University of Tübingen, Germany}
\affil[6]{Faculty of Health Sciences and Medicine, University of Lucerne, Switzerland}
\affil[{\Letter}]{\href{mailto:alceu.bissoto@unibe.ch}{alceu.bissoto@unibe.ch}, \href{mailto:lisa.koch@unibe.ch}{lisa.koch@unibe.ch}}

\date{}

\usepackage{lineno}
\begin{document}

\maketitle

\setcounter{footnote}{0}
\begin{abstract}

Machine learning (ML) models used in clinical workflows often perform systematically differently across patient groups, raising concerns about equity and patient safety. Traditional subgroup analyses can expose such disparities only when rich metadata (e.g., patient demographics) is available and captures the sources of performance variation, conditions rarely met in practice. \new{Algorithmic subgroup discovery methods offer a solution by identifying coherent groups from learned feature representations, but because true underlying stratifications are unknown, these methods are hard to evaluate.} As a result, subgroup discovery has not been systematically applied or assessed for performance monitoring. \new{Our main objective is to evaluate algorithmic subgroup discovery for performance monitoring, assessing the discovered subgroups' performance gaps and semantic coherence. Investigating cardiomegaly classification in a dataset of 223,228 chest X-rays, we showed that the discovered subgroups exhibited larger performance gaps (accuracies of 0.61--0.90) than traditional metadata-based analysis (0.80--0.94) and were frequently interpretable using patient, acquisition, and model metadata. As a proof of concept, we also showed that annotations from learned subgroups can be used to mitigate the biases they expose}. We conclude that subgroup discovery is a practical and effective tool to enhance the performance validation and monitoring of ML systems in healthcare.

\end{abstract}

\section*{Introduction}

Machine learning (ML) algorithms are implemented and commercialised as medical artificial intelligence (AI) devices at an unprecedented pace. In the US, for example, the FDA had approved 1,247 AI-assisted medical devices by May 2025, with 523 of these approvals occurring within the last 24 months~\cite{fda2025}. This growth reflects the vast potential of AI for patient care. 
However, research shows that even the most capable AI models can reinforce societal biases and fail unpredictably in clinical settings~\cite{zack2024assessing}. Consequently, regulatory agencies worldwide are tightening policies governing the development and deployment of medical AI to enhance its fairness and robustness. 

A critical concern is that ML models may perform systematically differently across patient subgroups \cite{seyyed2021underdiagnosis,christodoulou2024confidence,koch2024postmarket,oakden2020hiddenStratification}. This has hampered past attempts at safely deploying medical AI, particularly in underserved populations~\cite{seyyed2021underdiagnosis,christodoulou2024confidence}.  
Model performance can depend on many factors \cite{finlayson2021ClinicianDatasetShift}, including patient attributes (e.g., sex, age, ethnicity) and image characteristics (e.g., image quality, artifacts, device manufacturer). Chest X-rays model's performances, for example, were shown to depend on biological sex~\cite{larrazabal2020gender,weng2023sex},  support devices such as tracheotomy tubes and chest drains~\cite{jimenez2023detecting,olesen2024slicing}, artifacts (e.g., laterality markers, highlight arrows)~\cite{degrave2021ai}, and projection views~\cite{degrave2021ai} (anterior-posterior (AP) and posterior-anterior (PA)).
Not understanding when and why an ML model systematically fails poses an avoidable danger to patients and delays the development of more robust models. Furthermore, systematic hidden model failures may violate regulators' demands for safety and effectiveness for all intended use populations of a medical device.

These challenges are central to ensuring the safe and equitable application of AI in healthcare. Recent guidelines for trustworthy medical AI development, reporting, and deployment increasingly stress the importance of detailed subgroup analyses (e.g., TRIPOD+AI~\cite{collins2024tripod+}) and systematic bias assessments (e.g., FUTURE-AI~\cite{lekadir2025future}, PROBAST+AI~\cite{moons2025probast}). In response, current frameworks call on AI developers and dataset curators to explicitly identify and annotate potential sources of bias throughout the model development pipeline.
Together with clinical metadata, such as patient attributes and data-specific characteristics, the annotation can then be used to define patient subgroups to assess group fairness and evaluate model robustness. In general, the richer the available annotations, the more precisely relevant subgroups can be defined, increasing the chance of uncovering meaningful performance gaps. 

% 4. Limitations of subgroup analysis

However, acquiring such rich metadata is inherently challenging: dataset collection is both time- and personnel-intensive, and privacy concerns can limit the scope of metadata collection. In practice, this can lead to severe reporting gaps, as exposed by a recent scoping review of FDA-approved medical AI devices~\cite{muralidharan2024scoping}. According to their analysis, less than 4\% of approved devices provide information on race or ethnicity, less than 1\% report on socioeconomic factors, and less than 19\% report on age subgroups~\cite{muralidharan2024scoping}.

Even under ideal conditions, where multiple metadata attributes are available for a large, heterogeneous population, metadata may only partially align with the major hidden factors of variation that affect ML model performance. As a result, traditional metadata-defined subgroups may fail to reveal relevant performance gaps. Metadata annotations often reflect concepts that are meaningful for human interpretation and categorization, yet may be irrelevant to how machine learning models stratify data to make predictions. 
For example, stratifying a chest X-ray dataset by patient sex may be of limited value for auditing performance, since observed differences are often driven by correlated factors such as the presence of support devices (e.g., tubes and lines) rather than sex itself~\cite{weng2023sex, olesen2024slicing}. In contrast, stratifying by projection (AP vs PA) is crucial. Standard radiology interpretation emphasizes that the PA view is preferred for assessing cardiomegaly because the AP view magnifies the cardiac silhouette~\cite{raoof2012interpretation}. Accordingly, deep learning models perform better on PA studies than on AP studies for cardiomegaly-related labels (see Table~\ref{tab:subgroup-performance}).
Unrecognized subsets of data are referred to as \emph{hidden stratifications}. As we will demonstrate in this study, even in one of the most comprehensively annotated large-scale datasets available, CheXpertPlus~\cite{chambon2024chexpert}, which contains over 80 annotated factors including patient socio-demographics, comorbidities, and image characteristics for more than 200{,}000 samples, hidden stratifications remain prevalent. While manual inspection of model errors can sometimes reveal these issues, it is generally infeasible at scale and particularly challenging in medical imaging, where relevant differences across samples are often subtle.

Algorithmic subgroup discovery~\cite{eyuboglu2022domino,subbaswamy2024data} appears as a potential solution for this problem, as it can learn data stratifications that are relevant to a specific ML prediction model and dataset. The subgroup divisions are found by training unsupervised clustering methods (e.g., mixture models, k-means) on learned feature representations of the data.
Aside from the predictions from the investigated model, subgroup discovery methods also leverage complementary knowledge from models trained with different objectives (e.g., multi-modal, generative) or with more data. These models have been used so far only on toy datasets with predefined biases for easier evaluation.
We believe they might also be useful for performance evaluation and model surveillance, but they have yet to be proven effective in real-world medical contexts. The main challenge of using subgroup discovery techniques is how to validate the discovered subgroups, as hidden stratifications by definition lack ground-truth labels for evaluation. 

\begin{figure*}[t]
    \centering
    \includegraphics[]{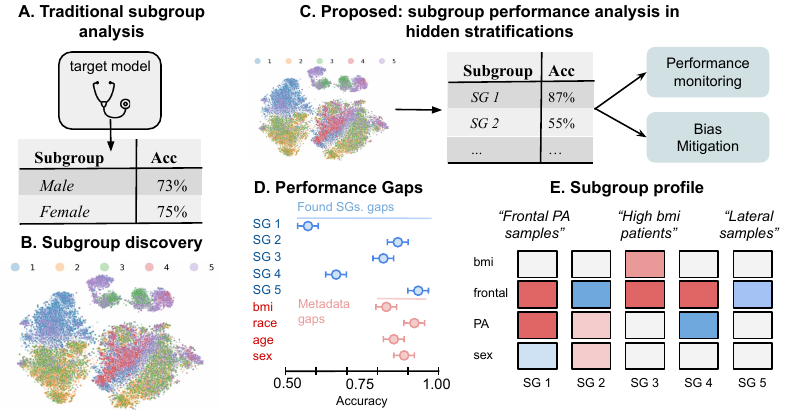}
    \caption{\new{From traditional subgroup analysis to algorithmic subgroup discovery for performance monitoring.} (a) Traditional subgroup analysis detects disparate patient outcomes, but it is limited to annotated metadata. (b) Subgroup discovery reveals hidden stratifications but lacks performance validation. (c) We bridge this gap by applying subgroup discovery for performance analysis in both (d) controlled synthetic settings and (e) real-world scenarios with unknown subgroups.
    }
    \label{fig:graphical-abstract}
\end{figure*}

% 6. This paper
% We need a stronger opening to make it clear what we want to do. Why is this important?
In this article, we thoroughly investigate algorithmic subgroup discovery\new{~\cite{eyuboglu2022domino,bissoto2025subgroup}} for detailed model performance evaluation (Fig.~\ref{fig:graphical-abstract}). For subgroup discovery techniques to be effective, they should satisfy two key criteria: (1) they must reveal substantial performance disparities across groups, and (2) the identified subgroups should exhibit semantic coherence. \new{That is, subgroup members should exhibit interpretable characteristics, whether related to patient (demographics, comorbidities), acquisition, or model factors. We define our research questions as:}

\new{
\begin{itemize}
    \item Do algorithmic subgroup discovery systematically expose performance gaps across different metrics? 
    \item Are the discovered subgroups semantically coherent, and can we associate them with existing metadata?
    \item Can the discovered subgroups be leveraged to mitigate the performance disparities they expose?
\end{itemize}
}
We employ a richly annotated real-world radiology dataset of chest x-ray images for a cardiomegaly-versus-all task to support a comprehensive evaluation. Using this dataset, we first contrast performance gaps identified through metadata-defined subgroups with those revealed by subgroup discovery.
We then characterize the discovered subgroups with our proposed method for creating metadata profiles, i.e. sets of metadata attributes that explain the similarities among samples within the same subgroup.

We argue that subgroup discovery can be an effective and easily implemented tool to enhance the performance validation and monitoring of ML systems in medicine. In this article, we expand on our prior work \cite{bissoto2025subgroup}, where we demonstrated preliminary evidence that algorithmic subgroup discovery can expose systematic performance gaps of medical ML models. Specifically, we propose a way to characterise found subgroups using existing metadata, which we call subgroup profiles. Moreover, we report additional metrics beyond accuracy and perform a statistical analysis of our results.
Overall, our main contributions are:
    \begin{itemize}
        \item We provide the first comprehensive evidence that subgroup discovery can systematically expose performance gaps in medical imaging, identifying meaningful subgroups in both synthetic and real-world settings.
        \item We demonstrate that discovered subgroups exhibit significantly larger performance disparities than conventional demographic metadata, revealing critical gaps missed by traditional fairness auditing.
        \item We show that discovered subgroups are cohesive \new{and that many exhibit statistically significant associations with annotated metadata}.
        \item We demonstrate that discovered subgroups can be effective for mitigating biases.
        
    \end{itemize}

\section*{Results}

We trained a deep learning model for cardiomegaly classification on CheXpertPlus~\cite{chambon2024chexpert}, a real-world chest X-ray dataset richly annotated with metadata, including patient demographics, acquisition details, and other factors (see first two columns of Table~\ref{tab:trad} for summary statistics).  \new{Our training, validation, and test sets follow an 80/10/10 split, with 178,684 / 22,263 / 22,281 images, respectively.} The model achieved a balanced accuracy of 0.70 in cardiomegaly classification and an area under the receiver-operating characteristic curve (AUROC) of 0.85 on the test set. Traditional subgroup analysis (metadata-based) revealed slight performance disparities between subgroups (see Table~\ref{tab:trad}). 

Then, we applied a simplified version \cite{bissoto2025subgroup} of a state-of-the-art subgroup discovery method~\cite{eyuboglu2022domino} to discover subgroups algorithmically from learned feature representations (see details in the Methods Section). %~\ref{sec:methods}). 
In short, the method used the cardiomegaly classification model's predictions and feature representations obtained from an external non-medical visual-language model \cite{clip}, and applied clustering to identify relevant subgroups. 
This resulted in 15 discovered subgroups \new{(denoted A0--A14)}, each containing between 223 and 4,354 images (blue bars in Fig.~\ref{fig:l3perfs}\,\textbf{A}). The prevalence of cardiomegaly varied across the subgroups between 8\% and 61\% (light blue bars in Fig.~\ref{fig:l3perfs}\,\textbf{B}). In contrast, traditional metadata contained between 711 and 18,894 samples (red bars in Fig.~\ref{fig:l3perfs}\,\textbf{A}), and cardiomegaly prevalence varied across metadata subgroups between 10\% and 24\% (light red bars in Fig.~\ref{fig:l3perfs}\,\textbf{B}). We report cardiomegaly prevalence because it highlights heterogeneity across subgroups with respect to the target pathology and because variation in prevalence directly affects measured performance, particularly given that specificity is typically higher than sensitivity.

\begin{table}[tb]
\newcolumntype{R}{>{\raggedleft\arraybackslash}X}
\newcolumntype{Y}{>{\arraybackslash}p{5.5cm}}
\centering
\small
\setlength{\tabcolsep}{4pt}
\renewcommand{\arraystretch}{1.2}
\begin{tabularx}{\textwidth}{@{} >{\arraybackslash}Y R N N N N @{}}
\toprule
\rowcolor{gray!20}
\textbf{Metadata Subgroups} & \textbf{Sample Size} &\textbf{Accuracy} & \textbf{Sensitivity} & \textbf{Specificity} & \textbf{AUROC} \\
\midrule
Full dataset & 22,281 & \numpm{0.872}{0.005} & \numpm{0.471}{0.018} & \numpm{0.936}{0.004} & \numpm{0.856}{0.007} \\
\hline
Sex: Male  & 13,134                  & \numpm{0.875}{0.006} & \numpm{0.479}{0.023} & \numpm{0.938}{0.004} & \numpm{0.859}{0.009} \\
Sex: Female & 9,127                  & \numpm{0.866}{0.007} & \numpm{0.459}{0.028} & \numpm{0.932}{0.006} & \numpm{0.850}{0.011} \\
\hline
Age: 0--60 & 10,032              & \numpm{0.905}{0.006} & \numpm{0.510}{0.030} & \numpm{0.952}{0.005} & \numpm{0.880}{0.011} \\
Age: 60+  & 12,229               & \numpm{0.844}{0.007} & \numpm{0.449}{0.022} & \numpm{0.921}{0.005} & \numpm{0.832}{0.010} \\
\hline
Race: White  & 12,635                 & \numpm{0.870}{0.006} & \numpm{0.430}{0.024} & \numpm{0.936}{0.005} & \numpm{0.845}{0.010} \\
Race: Black  & 1,164                 & \numpm{0.830}{0.022} & \numpm{0.598}{0.060} & \numpm{0.894}{0.020} & \numpm{0.858}{0.025} \\
Race: Asian   & 2,418                & \numpm{0.861}{0.014} & \numpm{0.479}{0.050} & \numpm{0.933}{0.011} & \numpm{0.861}{0.020} \\
\hline
Recent BMI: Normal (18.5 - 25)& 6,492           & \numpm{0.874}{0.008} & \numpm{0.428}{0.034} & \numpm{0.937}{0.006} & \numpm{0.852}{0.013} \\
Recent BMI: Morbidly Obese ($>$40)& 711    & \numpm{0.809}{0.029} & \numpm{0.465}{0.077} & \numpm{0.906}{0.025} & \numpm{0.830}{0.034} \\
\hline
Projection: PA  & 2,960                  & \numpm{0.889}{0.012} & \numpm{0.589}{0.050} & \numpm{0.932}{0.010} & \numpm{0.897}{0.016} \\
Projection: AP     & 16,032               & \numpm{0.869}{0.006} & \numpm{0.455}{0.021} & \numpm{0.936}{0.004} & \numpm{0.851}{0.009} \\
Support Devices: Positive & 11,501     & \numpm{0.875}{0.006} & \numpm{0.434}{0.025} & \numpm{0.943}{0.005} & \numpm{0.852}{0.010} \\
%Pacemaker: Positive     & 202             & \numpm{0.708}{0.062} & \numpm{0.568}{0.098} & \numpm{0.832}{0.071} & \numpm{0.789}{0.060} \\
View: Frontal  & 18,894   & \numpm{0.872}{0.005} & \numpm{0.474}{0.019} & \numpm{0.935}{0.004} & \numpm{0.859}{0.007} \\
View: Lateral  & 3,287   & \numpm{0.871}{0.012} & \numpm{0.452}{0.045} & \numpm{0.939}{0.009} & \numpm{0.835}{0.020} \\
Pneumothorax: Positive  & 1,696      & \numpm{0.945}{0.011} & \numpm{0.390}{0.247} & \numpm{1.000}{0.015} & \numpm{0.914}{0.082} \\
Edema: Positive & 5,454 & \numpm{0.800}{0.011} & \numpm{0.471}{0.027} & \numpm{0.095}{0.009} & \numpm{0.819}{0.013} \\
Lung Lesion: Positive & 908 & \numpm{0.924}{0.017} & \numpm{0.443}{0.110} & \numpm{0.032}{0.012} & \numpm{0.877}{0.037} \\
\bottomrule
\end{tabularx}
\caption{Traditional subgroup analysis. Subgroup performance metrics reported as value $\pm$ half 95\% CI. PA = posterior-to-anterior, AP = anterior-to-posterior.}
\label{tab:trad}
\end{table}

\subsection*{Subgroup discovery exposes higher performance gaps than traditional subgroup analysis}

\begin{figure*}[t]
    \centering
    \includegraphics{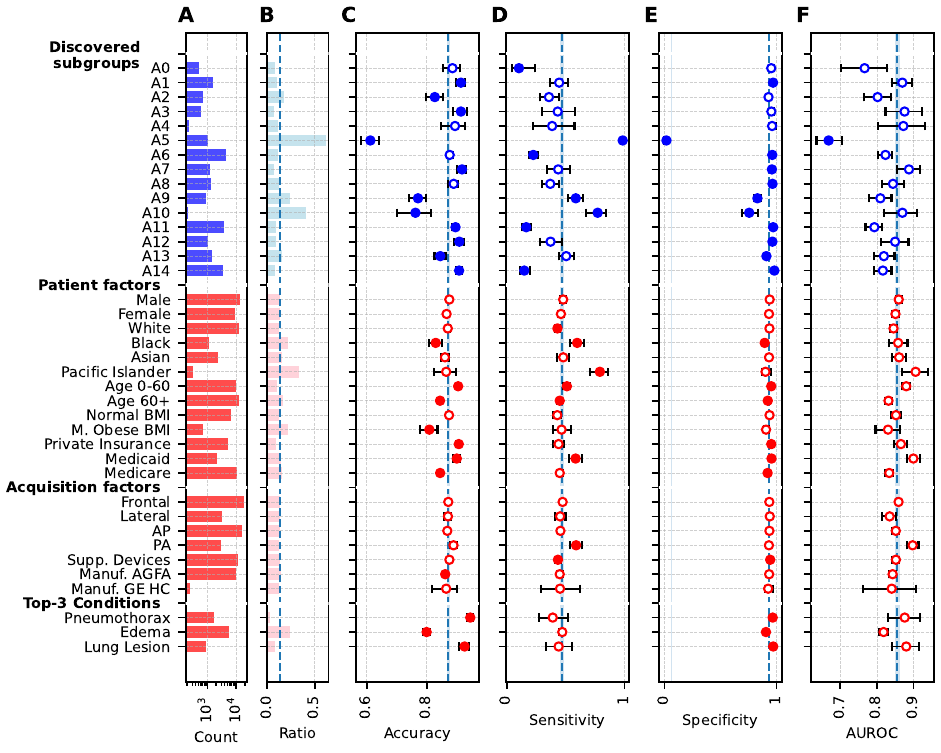}
    \caption{Performances of discovered subgroups (blue) and traditional metadata subgroups (red). Filled points indicate statistically significant performance disparities from its complement. For each group, we show their size (A), cardiomegaly class ratio (B), and their performance in prevalence-dependent (C) and independent (D-F) metrics.}
    \label{fig:l3perfs}
\end{figure*}

Discovered subgroups showed larger performance disparities than traditional metadata-defined subgroups (Fig.~\ref{fig:l3perfs}, panels \textbf{C–F}). Accuracy varied substantially across discovered subgroups, ranging from 0.61 to 0.92, which may partly reflect differences in class prevalence (0.08–0.61). However, we also observed statistically significant differences in prevalence-independent metrics: sensitivity ranged from 0.10 to 0.98, and AUROC from 0.67 to 0.89. \new{This shows that prevalence alone is insufficient to justify the disparities observed among the discovered subgroups.} Overall, sensitivity differed significantly in 7 of 15 subgroups, specificity in 10 of 15, and accuracy in 11 of 15 (see details about the statistical study in the Methods Section). In contrast, metadata-defined subgroups showed smaller but still significant gaps, with accuracy ranging from 0.80 to 0.94, sensitivity from 0.39 to 0.60, and AUROC from 0.82 to 0.88. Across these, sensitivity differed significantly in 6 of 17 subgroups and accuracy in 8 of 17.

The largest disparities were concentrated in a few specific subgroups. Among discovered subgroups, groups \new{A}5 and \new{A}10 showed the highest sensitivity (0.984 and 0.772, respectively) and the lowest specificity (0.017 and 0.756). The pronounced sensitivity–specificity imbalance in subgroup \new{A}5 reflects a strong bias toward positive cardiomegaly predictions, resulting in the only significant AUC disparity (0.188) across all experiments and the largest accuracy gap (0.260). Subgroup \new{A}9, despite a relatively low prevalence of positive samples, also showed a marked reduction in accuracy of 0.101. Among metadata-defined subgroups, Black patients exhibited the largest disparity, with sensitivity increased by 0.13, specificity decreased by 0.05, and accuracy decreased by 0.03 relative to the average. Regarding acquisition factors, posteroanterior (PA) samples showed the largest disparities, with sensitivity increased by 0.12 and accuracy by 0.03.

\subsection*{Discovered subgroups are diverse and \new{often align with clinical or acquisition-related concepts}}

\begin{figure*}[h]
    \centering
    \includegraphics{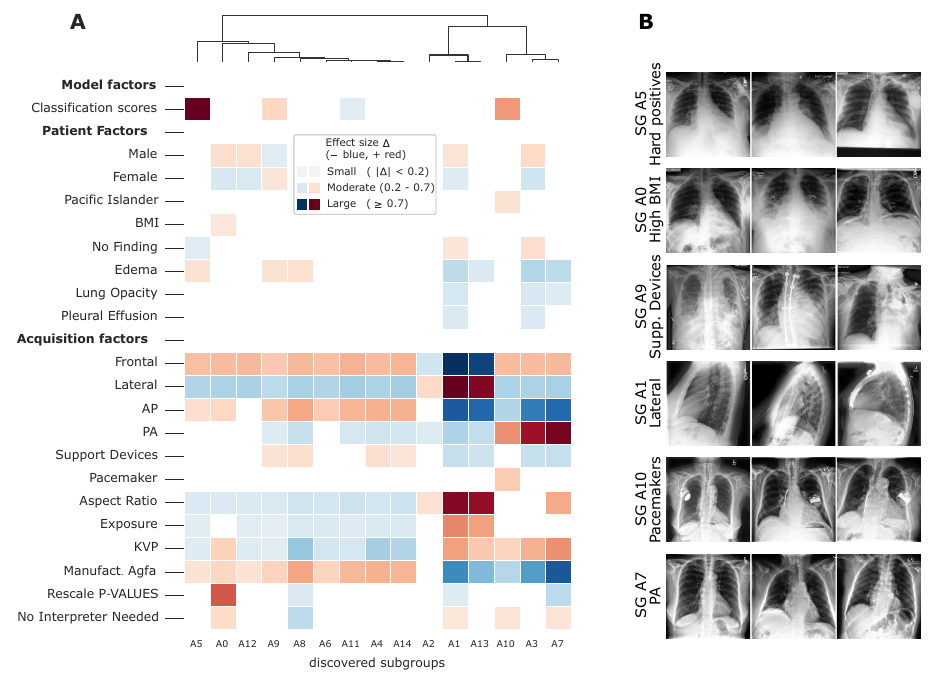}
    \caption{\new{Characteristics of discovered subgroups.} (A) Discovered subgroup's metadata profile. Each subgroup (column) is characterized by a set of metadata (row), and its effect (cells) is measured by Cohen's h for categorical values and Cohen's d for continuous ones. (B) This characterization enables us to understand the subgroups, identify key characteristics important to the classification models, and create performance gaps. For a few selected subgroups, we show four randomly selected images from this group, and their most distinctive characteristic.}
    \label{fig:heatmap}
\end{figure*}

We next verified whether the discovered subgroups could be characterised with semantic concepts. For this, we used the rich metadata available in the CheXpertPlus~\cite{chambon2024chexpert} dataset, and created metadata profiles for each discovered subgroup by measuring the association between the metadata attributes and each subgroup (see the Methods section for details). The resulting metadata profiles can be analysed as heatmaps for each discovered subgroup (columns in Fig.~\ref{fig:heatmap}\textbf{A}), with bold colours indicating a strong association between a subgroup and a metadata attribute. \new{We divide the subgroups into three major categories: patient, acquisition, and model factors.}

% Table

\begin{table}[tb]
\centering
\small
\setlength{\tabcolsep}{1pt}
\renewcommand{\arraystretch}{1.2}
\begin{tabularx}{\textwidth}{@{} >{\centering\arraybackslash}p{0.9cm} >{\raggedright\arraybackslash}p{1.6cm} Y N N N N @{}}
\toprule
\rowcolor{gray!20}
\new{\textbf{SG}} & \new{\textbf{Category}} & \textbf{Description} & \textbf{Accuracy} & \textbf{Sensibility} & \textbf{Specificity} & \textbf{AUROC} \\
\midrule
A5  & \noP~\noA~\catM & \new{Positive preds}                 & \numpm{0.612}{0.030} & \numpm{0.984}{0.010} & \numpm{0.017}{0.014} & \numpm{0.668}{0.034} \\
A0  & \catP & \new{High BMI, males}                                & \numpm{0.886}{0.028} & \numpm{0.103}{0.097} & \numpm{0.952}{0.019} & \numpm{0.766}{0.063} \\
A12 & \catP &\new{Males, unknown}                                        & \numpm{0.909}{0.017} & \numpm{0.371}{0.094} & \numpm{0.963}{0.012} & \numpm{0.849}{0.038} \\
A9  & \catP~\catA~\catM & \new{Females, positive preds, support devices}
                                                   & \numpm{0.771}{0.028} & \numpm{0.586}{0.066} & \numpm{0.829}{0.029} & \numpm{0.809}{0.031} \\
A8  & \noP~\catA & \new{Support devices, interpreter needed}        & \numpm{0.890}{0.018} & \numpm{0.369}{0.074} & \numpm{0.964}{0.011} & \numpm{0.844}{0.029} \\
A6  & -- & \new{Unknown}                  & \numpm{0.876}{0.010} & \numpm{0.224}{0.037} & \numpm{0.962}{0.006} & \numpm{0.823}{0.018} \\
A11 & \noP~\noA~\catM & \new{Negative preds, unknown}                          & \numpm{0.896}{0.010} & \numpm{0.165}{0.038} & \numpm{0.970}{0.005} & \numpm{0.793}{0.023} \\
A4  & \noP~\catA & \new{Support devices, unknown}                               & \numpm{0.894}{0.040} & \numpm{0.385}{0.175} & \numpm{0.960}{0.028} & \numpm{0.872}{0.064} \\
A14 & \noP~\catA & \new{Support devices, unknown}                               & \numpm{0.908}{0.010} & \numpm{0.148}{0.041} & \numpm{0.980}{0.005} & \numpm{0.816}{0.024} \\
A2  & \noP~\catA & \new{Lateral samples, unknown} & \numpm{0.827}{0.028} & \numpm{0.358}{0.083} & \numpm{0.928}{0.022} & \numpm{0.802}{0.037} \\
A1  & \catP~\catA  &  \new{Lateral images, healthy, males} & \numpm{0.914}{0.014} & \numpm{0.444}{0.076} & \numpm{0.969}{0.010} & \numpm{0.869}{0.029} \\
A13 & \noP~\catA &  \new{Lateral images}                   & \numpm{0.845}{0.019} & \numpm{0.504}{0.063} & \numpm{0.909}{0.016} & \numpm{0.819}{0.027} \\
A10 & \catP~\catA~\catM &  \new{Positive preds, PA, pacemaker, Pacific Islander}
                                                   & \numpm{0.762}{0.055} & \numpm{0.772}{0.085} & \numpm{0.756}{0.073} & \numpm{0.869}{0.045} \\
A3  & \catP~\catA & \new{Healthy males, PA}                & \numpm{0.914}{0.023} & \numpm{0.432}{0.140} & \numpm{0.953}{0.018} & \numpm{0.875}{0.050} \\
A7  & \noP~\catA & \new{PA, unknown}                             & \numpm{0.918}{0.016} & \numpm{0.436}{0.099} & \numpm{0.958}{0.011} & \numpm{0.887}{0.031} \\
\bottomrule
 \end{tabularx}
\caption{Performance by discovered subgroups.  \new{Subgroups are categorized into patient (\catP), acquisition (\catA), and model (\catM) factors. }Values are mean~$\pm$~95\% confidence-interval half-width. PA = posterior-to-anterior, AP = anterior-to-posterior.}
\label{tab:subgroup-performance}
\end{table}

\new{Patient factors include metadata commonly used in traditional subgroup analyses, such as patient demographics and comorbidities.} These appear relevant to some subgroups but only to a limited extent. The only ethnicity that showed a significant association was ``Pacific Islander'' on subgroup 10. Males and females showed a mild influence on five (one-third) groups. 
Medical findings, such as co-occurring lung lesions, showed mild to moderate impact. Of the 12 co-morbidities (excluding No Findings) annotated in CheXpert, only 3 showed significant associations, with Edema being the most common, appearing in 7 groups.  

\new{Acquisition factors were the most strongly associated with our discovered subgroups. 
Image characteristics such as Frontal/Lateral and AP/PA strongly influenced subgroups, as different views and projections consistently clustered together and were rarely assigned to mixed subgroups. Since these were often binary complementary values (e.g., frontal vs. lateral, AP vs. PA), a high positive association on one led to a high negative association on the other.} 
%Other factors exhibited a significant, but milder effect.
External objects, such as support devices, were significantly associated with some of the identified subgroups. Since these objects appeared highly relevant to our investigated models, we included annotations of pacemakers in our analysis. For this, we used publicly available annotations~\cite{weng2024fast}, which covered 200 samples of our test set. Even if annotated for only a small subset of the dataset, pacemakers became a defining attribute in one of our worst-performing subgroups (\new{SG A}10).

\new{Model factors, such as the classification scores, were a defining attribute, particularly for subgroups 5 and 10, the worst-performing subgroups. That shows that these subgroups contained samples with systematically higher predicted model scores (positive class predictions) than the other subgroups.} 

The metadata profiles described above and shown in Fig.~\ref{fig:heatmap} allow an augmented and enriched version of traditional subgroup analysis. Based on the obtained metadata profiles for the discovered subgroups, we created an enriched subgroup analysis table (Table~\ref{tab:subgroup-performance}) to showcase how subgroup discovery could be implemented to complement traditional subgroup analysis for model validation. As an important limitation, the discovered subgroups could only be explained with metadata that were available during analysis. While we could show that discovered subgroups were strongly linked with some semantic features, some factors remained unknown.
For example, subgroups \new{A}1 and \new{A}13 had similar profiles but differed by 6 percentage points in accuracy, and subgroups \new{A}4 and \new{A}14 differed by 24 percentage points in sensitivity. Subgroup discovery as a tool can therefore expose differences beyond available metadata, and could be used to trigger further investigations by domain experts to identify the missing defining attributes of these subgroups.

\subsection*{Use case: subgroup annotations for improving model robustness}

Here, we present model debiasing as a downstream use case of subgroup discovery. Typically, model debiasing and domain generalisation rely on subgroup annotations (e.g., image domain, patient sex, presence of artifacts). The subgroup labels are then used during model training to improve worst-group performance and thus improve model robustness and fairness.
Subgroup discovery can automatically generate such annotations and identify underperforming subgroups and potential biases, supporting subsequent model improvement through debiasing. 

% Here, we focused on the second case. 
Manual inspection revealed a consistent presence of pacemakers in \new{SG A1}0, which was also the second-worst-performing group. This suggests that the model relied on pacemaker features, leading to reduced performance. 
While pacemakers were not the only defining attribute of this low-performing subgroup, nor did the subgroup include all pacemaker samples, we hypothesized that improving robustness with respect to Subgroup \new{A}10 could mitigate performance drops in the pacemaker cohort.
If effective, subgroup discovery could replace much of the manual annotation needed to identify artifacts such as pacemakers.

We used groupDRO \cite{sagawa2019distributionally}, a standard domain generalization approach that minimizes the worst subgroup error. We created an annotation indicating whether a sample belonged to Subgroup \new{A}10. For evaluation, we used a publicly available partial annotation of pacemakers \cite{weng2024fast}, which covered 200 samples in our test set. Although small, this subset reflects real-world conditions, as such annotations are costly to obtain. After hyperparameter tuning and training, the groupDRO model improved balanced accuracy on pacemaker cases by 3.2 percentage points (from 70.0 to 73.2). All training and validation, including hyperparameter tuning, used only the subgroup-based annotations, while the pacemaker reference labels were reserved for final evaluation on the test set. 

% Synthetic case. Add a brief explanation of the methodology.
\subsection*{Subgroup discovery passes sanity checks on synthetic data} \label{sec:results_known_biases}

In real-world medical imaging, detailed metadata are often scarce, and when available, they may not reflect the features that most influence the model’s predictions. Subgroup discovery is designed for exactly these situations, where few or no annotations exist. However, this makes its evaluation challenging, since we cannot rely on supervised ground truth to assess the discovered subgroups.

To address this, we created a synthetic version of CheXpert that simulates performance disparities while providing access to known, ground-truth subgroups. Following previous work~\cite{sun2023right}, we added two clinically inspired artifacts correlated with the positive disease label: vertical lines of hyperintense signal and a hospital tag in the bottom-left corner. Each artifact combination defines a ground-truth subgroup (no artifacts, only vertical line, only tag, both). We trained the classification model and applied subgroup discovery on this biased dataset, without using any artifact information. Evaluation was then performed on an unbiased test set, in which artifacts were equally likely to appear in both positive and negative samples. This experiment tests whether subgroup discovery methods can (1) identify and isolate these subgroups and (2) capture the performance disparities introduced by the artifacts.

%\paragraph{Isolating ground truth subgroups.}

\begin{figure}[h]
    \centering
    \includegraphics{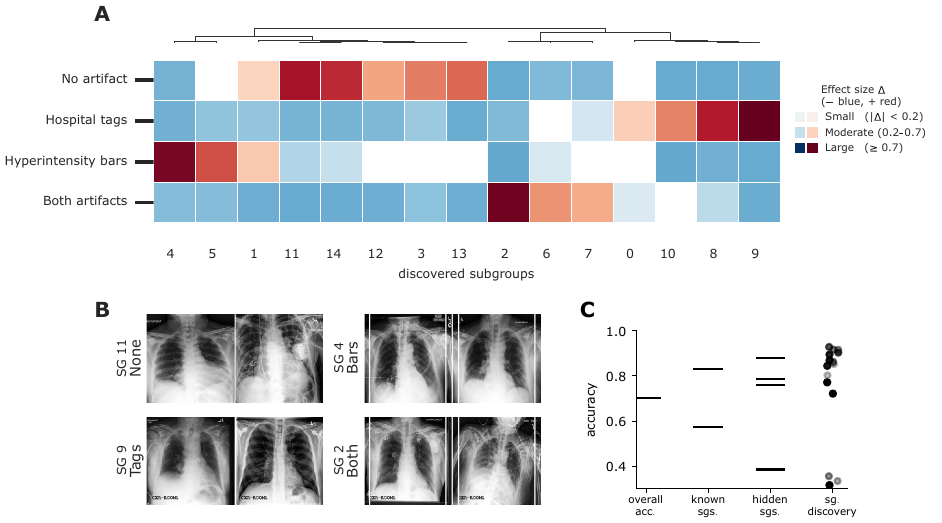}

    \caption{\new{Sanity checks of subgroup discovery.} (A) Effects of known simulated artifacts. Subgroup discovery successfully dedicates subgroups for each possible combination of the presence of the two artifacts. (B) Random samples from the found subgroups. (C) Detailed subgroup accuracies, reproduced from \cite{bissoto2025subgroup}. Darker dots represent found subgroups with higher effects. These capture the true performance gap characterized by hidden subgroups, which are overlooked by traditional subgroup analysis based on a single artifact (known subgroups), and by overall performance.
    }
    \label{fig:l2profile}
\end{figure}

In Fig.~\ref{fig:l2profile}\textbf{A}, we showed that subgroup discovery successfully isolated all combinations of artifacts. The artifact concepts were only mildly mixed across subgroups (Fig.~\ref{fig:l2profile}\textbf{B}), because these are highly impactful features by design (high-bias scenario). As this synthetic data is an altered version of CheXpert, it also retains the dataset's original characteristics, such as frontal/lateral and AP/PA, which might still affect subgroups' composition. Here we had access to the metadata inherited from CheXpert, as shown in previous results, but we omitted them for clarity.

%\paragraph{Matching performance disparities}

\new{In Fig.\,\ref{fig:l2profile}\textbf{C}\, we showed that subgroup discovery captured the ground-truth (based on the introduced artifacts) hidden subgroup performances.}
We contrasted the discovered subgroups' performances with those calculated having access to none (``overall acc.'' column), one (``known sgs.''), or both artifacts (``hidden sgs.''). Having none or one artifact annotated represents the more realistic scenario, as biases are often not annotated. Having all artifacts annotated serves as ground truth for the discovered subgroups. When only one artifact is known, subgroup analysis based on the known artifact unsurprisingly revealed increasing performance gaps, but missed the much larger, hidden performance gaps caused by the second artifact.

\subsection*{Feature extractors trained on natural images are sufficient for exposing meaningful performance gaps}

\begin{table}[tbh]
\centering
\small
\setlength{\tabcolsep}{1pt}
\renewcommand{\arraystretch}{1.2}
\begin{tabularx}{\textwidth}{@{} >{\centering\arraybackslash}p{0.9cm} >{\raggedright\arraybackslash}p{1.6cm} Y N N N N @{}}
\toprule
\rowcolor{gray!20}
\new{\textbf{SG}} & \new{\textbf{Category}} & \textbf{Description} & \textbf{Accuracy} & \textbf{Sensibility} & \textbf{Specificity} & \textbf{AUROC} \\
\midrule
B4  & \noP~\catA & \new{Lateral} & \numpm{0.804}{0.033} & \numpm{0.432}{0.091} & \numpm{0.896}{0.029} & \numpm{0.810}{0.044} \\
B0  & \catP~\catA & Young, healthy, lateral & \numpm{0.919}{0.014} & \numpm{0.317}{0.075} & \numpm{0.977}{0.008} & \numpm{0.839}{0.035} \\
B1  & \noP~\catA & \new{Lateral, no support devices} & \numpm{0.839}{0.020} & \numpm{0.528}{0.064} & \numpm{0.904}{0.018} & \numpm{0.814}{0.030} \\
B6  & \catP~\catA~\catM & \new{PA, healthy, negative preds} & \numpm{0.944}{0.013} & \numpm{0.258}{0.104} & \numpm{0.979}{0.008} & \numpm{0.855}{0.043} \\
B2  & \noP~\noA~\catM & Positive preds & \numpm{0.608}{0.026} & \numpm{0.961}{0.014} & \numpm{0.104}{0.026} & \numpm{0.668}{0.030} \\
B9  & \noP~\catA~\catM & \new{Pacemaker, PA, positive preds} & \numpm{0.730}{0.042} & \numpm{0.775}{0.060} & \numpm{0.698}{0.056} & \numpm{0.815}{0.038} \\
B8  & \catP~\noA~\catM & \new{Pleural effusion, consolidation, negative preds} & \numpm{0.930}{0.020} & \numpm{0.182}{0.113} & \numpm{0.988}{0.010} & \numpm{0.847}{0.060} \\
B13 & \catP~\noA~\catM & \new{Pleural effusion, negative preds} & \numpm{0.908}{0.021} & \numpm{0.107}{0.083} & \numpm{0.976}{0.012} & \numpm{0.795}{0.054} \\
B5  & \noP~\catA & PA samples, unknown & \numpm{0.870}{0.014} & \numpm{0.375}{0.055} & \numpm{0.936}{0.010} & \numpm{0.840}{0.021} \\
B14 & \catP & Older patients & \numpm{0.866}{0.018} & \numpm{0.167}{0.055} & \numpm{0.972}{0.010} & \numpm{0.790}{0.035} \\
B11 & \catP & Healthy, unknown & \numpm{0.908}{0.017} & \numpm{0.191}{0.081} & \numpm{0.971}{0.011} & \numpm{0.815}{0.047} \\
B3  & \catP~\catA & \new{Lung opacity, effusion, support devices} & \numpm{0.830}{0.028} & \numpm{0.276}{0.080} & \numpm{0.939}{0.020} & \numpm{0.790}{0.039} \\
B10 & \catP~\catA~\catM & \new{Lung opacity, consolidation, support devices, negative preds}& \numpm{0.899}{0.021} & \numpm{0.085}{0.067} & \numpm{0.980}{0.011} & \numpm{0.805}{0.048} \\
B7 & \catP~\noA~\catM  & Support devices, negative preds & \numpm{0.913}{0.009} & \numpm{0.118}{0.035} & \numpm{0.979}{0.004} & \numpm{0.810}{0.022} \\
B12 & \noP~\catA & Support devices, unknown & \numpm{0.890}{0.011} & \numpm{0.163}{0.039} & \numpm{0.974}{0.006} & \numpm{0.805}{0.022} \\
\bottomrule
\end{tabularx}
\caption{Discovered subgroups using BioMedCLIP at threshold 0.5. \new{Subgroups are categorized into patient (\catP), acquisition (\catA), and model (\catM) factors.} Metrics are reported as mean $\pm$ half 95\% CI.  PA = posterior-to-anterior.}
\label{tab:subgroups-cxr}
\end{table}

In the main results presented above, we used a pretrained natural image feature extractor \new{(CLIP~\cite{clip})} in the subgroup discovery step. Next, we investigated whether representations learned from biomedical data led to better stratification of disease-related features in medical images. For this, we used BiomedCLIP~\cite{zhang2024biomedclip}, a foundation model trained on a wide range of biomedical multimodal datasets, as a feature extractor for subgroup discovery.

Overall, BiomedCLIP and generalist CLIP led to similar performance disparities, but different metadata profiles (see Table~\ref{tab:subgroups-cxr} vs. Table~\ref{tab:subgroup-performance}).
The subgroups obtained with BiomedCLIP were more strongly associated with comorbidities such as lung diseases. For generalist CLIP, no subgroup was linked to lung diseases, but BiomedCLIP found positive associations in 4 out of 15 subgroups. Further, it identified a subgroup of older patients but missed links with BMI, potentially because age was more often mentioned in radiology reports used for training the BiomedCLIP foundation model. 
Despite these differences, our results show that even feature extractors trained on natural images can expose meaningful performance gaps in real-world medical data distributions. The different subgroup characteristics call for combining external models in future work to yield even more diverse subgroups.

\section*{Discussion}

Deep learning models frequently show systematic performance differences across patient populations, making careful subgroup analyses essential for uncovering these disparities. Traditional subgroup analysis relies on metadata such as sex or age to stratify performance, but these attributes often fail to align with the model’s underlying decision concepts. As a result, conventional approaches may miss important systematic differences. Subgroup discovery offers a way to identify more relevant groups, yet its evaluation remains difficult because ground-truth subgroup annotations are unavailable for real-world data.

We explored this challenging problem using a real-world chest X-ray dataset~\cite{chambon2024chexpert} chosen for its extensive metadata. Even under these unusually favourable conditions, we show that metadata alone does not adequately reveal meaningful subgroup disparities. In contrast, subgroup discovery uncovered groups exhibiting larger and statistically significant disparities across both prevalence-dependent metrics (e.g., accuracy) and prevalence-independent metrics (e.g., sensitivity, specificity, AUC).
\new{These findings extend prior work on subgroup discovery~\cite{eyuboglu2022domino,bissoto2025subgroup} by proposing a way to characterise found subgroups using existing metadata, which we call subgroup profiles, and by thoroughly evaluating the performance gaps beyond accuracy and performing a rigorous statistical analysis.} 

\new{Revealing performance disparities in high-performing models addresses only part of the challenge. It is equally important to show that the identified subgroups are internally coherent, that is, that their samples share a meaningful, potentially interpretable concept. To support interpretability, we examined whether specific metadata values were overrepresented within each subgroup relative to all remaining samples. 
Interpretability is a multifaceted concept~\cite{nauta2023anecdotal} involving content properties (fidelity to the model's internal behaviour), presentation properties (how explanations are communicated), and user-facing properties (whether explanations are practically relevant to the end user). Our pipeline contributes to each axis: the statistical analysis and sanity checks establish content fidelity; the metadata profiling translates subgroup structure into clinically legible terms; and the actionability analysis below assesses user-facing relevance to determine whether and how to act.}

\new{The EU AI Act~\cite{euaiact2025} operationalizes this final step for high-risk medical AI, requiring that systems enable deployers to "interpret outputs correctly and use the system appropriately" (Article 13), and establishing further obligations for data governance (Article 10), robustness (Article 15), human oversight (Article 14), and post-market surveillance (Article 72).}

%Our study revealed subgroups with strong associations to annotated metadata, including clinically relevant factors such as patient demographics and comorbidities, as well as imaging-driven factors such as acquisition parameters. Other subgroups could not be attributed to any annotated factor, or were distinguished primarily by patterns in the model's own outputs, such as systematically elevated prediction confidence. 
\new{Regardless of whether a subgroup can be fully characterised, the discovery of a performance disparity enables an immediate response: flagging the affected group for deferral to specialist review and for stratified performance monitoring, addressing human oversight and post-market surveillance obligations. For example, SG~A0, associated with high-BMI males, shows a sensitivity of 0.10 and a specificity of 0.95, indicating that positive cases in this group are rarely detected. A similar pattern appears in SG~B14, associated with older patients (sensitivity 0.17, specificity 0.97).  In both cases, patients with these characteristics can be flagged for manual assessment rather than relying on AI-based predictions, preventing potentially life-threatening missed diagnoses.}

\new{Beyond deferral, the recommended intervention depends on how a subgroup's characteristics interact with the classification target. This requires domain expertise and may involve both clinicians and computer scientists. We distinguish three scenarios. Firstly, the factors associated with a subgroup can be clinically relevant to the disease (e.g., patient BMI and age). In this case, subgroup-level model recalibration may be appropriate to improve the model and mitigate the performance gap. Secondly, the associated factor introduces systematic imaging distortions that compromise diagnostic accuracy, for instance, differences between AP and PA chest projections~\cite{raoof2012interpretation}. Here, an appropriate short-term response may be to flag the patient for additional imaging to reduce ambiguity. In the longer term, such findings motivate targeted robustness testing and corrections. Thirdly, the identified characteristic may confound the classifier, as with the pacemaker-enriched SG A10. Here, targeted domain adaptation using subgroup labels can mitigate the disparity~\cite{alloula2025subgroups}, as we demonstrated in our proof-of-concept experiment. In each case, the subgroup discovery can identify potential problem sources, and specialist assessment determines the appropriate response.}

\new{Some subgroups in our analyses could not be attributed to any annotated clinical, demographic, or acquisition factor. While metadata-based characterisation reduces the burden of manual interpretation and scales to large datasets, it is inherently bounded by what is annotated. For underperforming subgroups with no apparent clinical explanation, the appropriate response starts with a domain expert examining the flagged samples to determine whether they share a concept not captured by available metadata annotations. Once such a concept is identified, it may be linked to one of the scenarios and appropriate interventions described above. If no unifying concept emerges, as with SG A5, whose systematically inflated classifier confidence persists after controlling for all annotated factors, or SG A6, which shows no strong association with any specific characteristic, the discovered subgroups remain actionable: they provide a basis for deferral given their known elevated error rate, and they serve as challenging holdout subgroups for ongoing model validation.}

Several limitations should be considered. 
\new{The metadata-based characterisation, while extensive by the standards of publicly available datasets, is constrained by what was annotated. Factors not captured in the metadata (e.g., disease stratifications, patient positioning details) may drive subgroup structure that our profiling cannot expose. Finally, our analysis identified associations between subgroup membership and metadata factors, but it does not establish causal mechanisms for the observed performance gaps. }

We posit that the current study represents a significant advance in the ability of model developers and evaluators to better assess the safety and effectiveness of a model across a range of patient populations. Subgroup discovery can unveil ongoing problems and provide informative labels for model debiasing. 
We conclude that subgroup discovery should be highly relevant as a performance monitoring and reporting tool, and argue that it should accompany traditional subgroup analysis as an additional safeguard during real-world ML validation and deployment.
\section*{Methods} \label{sec:methods}

We seek subgroup divisions that expose large systematic performance gaps in a target classification model while preserving subgroup cohesion, so that the model's performance within a subgroup can be attributed to a shared characteristic. 
Metadata-based subgroups are inherently cohesive because a semantic concept, such as patient age or sex, defines the division. However, we hypothesise that these attributes do not adequately reflect the main factors that affect model performance, which often results in relatively small performance gaps (Fig.\,\ref{fig:graphical-abstract}\,\textbf{a}). Instead, we propose to use subgroup discovery techniques (Fig.\,\ref{fig:graphical-abstract}\,\textbf{b}) for subgroup performance analysis in hidden stratifications (Fig.\,\ref{fig:graphical-abstract}\,\textbf{c}). 

\new{First, we train the classification model (see ``Experimental setup'') for detecting cardiomegaly following the train/val/test splits. Then, we fit a simplified version~\cite{bissoto2025subgroup} of DOMINO~\cite{eyuboglu2022domino} for subgroup discovery on the validation set (see ``Preliminaries: subgroup discovery algorithm'') to identify latent subgroups. Finally, we predict the subgroup assignments of test samples for subsequent analysis.}

%We propose a two-tiered evaluation approach to tackle the difficult challenge of validating hidden subgroups.
We propose a strategy to evaluate subgroup discovery for performance analysis for the first time in a real-world medical data distribution (Fig.\,\ref{fig:graphical-abstract}\,\textbf{e}). \new{Here, we first calculate performance gaps using different metrics and performing statistical tests (see ``Statistical analysis of subgroup performance''). Then, we use existing metadata to characterize the discovered subgroups (see ``Statistical analysis of metadata profiles'').}
As a sanity check, we inject synthetic artifacts to create clinically-inspired subgroups \new{see ``Synthetic scenario with generated artifacts'')} for which ground-truth model performance is available (Fig.\,\ref{fig:graphical-abstract}\,\textbf{d}). 

\subsection*{Dataset description}\label{sec:datasets}
We selected a dataset that provides comprehensive metadata coverage.
CheXpert-Plus~\cite{chambon2024chexpert} is an extension of CheXpert~\cite{irvin2019chexpert} and provides metadata that allows for a challenging comparison to our discovered subgroups. The metadata includes patient demographics (e.g., sex, age), 
comorbidities (e.g., edema, fracture), 
and artifacts,
totalling 20 attributes. Our training, validation, and test sets follow an 80/10/10 division, with a total of 178,684 / 22,263 / 22,281 images~respectively.

\subsection*{Preliminaries: subgroup discovery algorithm}
We use DOMINO\,\cite{eyuboglu2022domino}, a simple yet effective approach for subgroup discovery. 
First, a feature representation $z(x)$ is extracted from each image $x$ using an external pretrained model such as CLIP \cite{clip}, followed by dimensionality reduction using principal component analysis. In addition, softmax predictions $\hat{y}(x)$ are obtained from the target classification model. 
While the model predictions encapsulate characteristics important for the classification task, the external model helps identify task-agnostic features such as artifacts. Next, the samples are clustered into subgroups $\mathcal{S}$ using a generalised Gaussian Mixture Model (GMM) by minimising the following objective (similar to \cite{eyuboglu2022domino}):
\begin{equation}
    \ell(\phi) \!=\! \!\!\!\!\!\!\!\sum_{i=1}^{n_{samples}}\!\!\!\!\! \log \! \sum_{j=1}^{|\mathcal{S}|} P_{\phi_S}(\mathcal{S}^{(j)}\! =\! 1) P_{\phi_Z}\!(Z\!\! =\!\! z(x_i)\! \mid \!\mathcal{S}^{(j)}\!\! =\!\! 1) P_{\phi_{\hat{Y}}}\!(\hat{Y}\!\! =\!\! \hat{y}(x_i) \!\mid\! \mathcal{S}^{(j)}\! \!=\!\! 1)^\gamma,
    \label{eq:domino}
\end{equation}
\noindent where $\gamma$ balances the influence of predicted labels $\hat{y}(x)$ and embeddings $z(x)$ in the slicing decision. 
In contrast to the original DOMINO \cite{eyuboglu2022domino}, we remove classification labels in the GMM, enabling subgroup discovery in post-deployment scenarios with unlabeled test sets.
We use explicit validation and test set splits, fitting DOMINO on the validation set and inferring subgroups on the test set.

\subsection*{Statistical analysis of subgroup performance} \label{sec:statistical_perf}

We computed subgroup performance metrics by estimating prevalence-dependent (accuracy), and independent metrics (sensitivity, specificity, and AUROC) within each predefined or discovered subgroup. We evaluated accuracy, sensitivity, and specificity at a fixed decision threshold of 0.5, and obtained their confidence intervals using Wilson's score intervals at the 95\% level. For AUROC, we report a 95\% bootstrap interval (2.5th and 97.5th percentiles) using 1,000 resamples drawn with replacement within each subgroup. Resamples that contained only one class were skipped.

To assess disparities, each subgroup was compared to its complement (all remaining examples not belonging to the subgroup). For sensitivity, specificity, and accuracy, we used a two-sample test for proportions by taking the difference in the rates, dividing that difference by its estimated standard error, and obtaining a z value that reflects how far apart the two rates are under the assumption that they are truly equal.
Analogously, for AUROC, we use DeLong's method for two independent samples. Since we conducted many subgroup comparisons, we applied a Bonferroni correction to the p-values to keep the false-positive rate below 0.05. We considered a difference significant when the corrected p-value was below this threshold.

\subsection*{Statistical analysis of metadata profiles}

To characterize the discovered subgroups with semantic information, we used the available metadata to create metadata profiles describing how each attribute relates to each subgroup. For each subgroup, we compared the distribution of each metadata attribute within the subgroup with that of the rest of the dataset.
For categorical attributes (e.g., sex, race, view position), we tested each category separately (e.g., sex = male) against the combined set of all other categories. Missing categorical values were recoded as an explicit “Unknown” category and included in the analysis. For CheXpert comorbidities, missing and negative labels were treated as absent, while uncertain labels were kept as a separate level. Differences in proportions were measured using Cohen’s h, where positive values indicate that a category is more common within the subgroup and negative values indicate it is less common. For continuous attributes (e.g., age, BMI), we computed Cohen’s d, the standardized mean difference between the samples within and outside the subgroup, with positive values indicating higher means within the subgroup.

Statistical significance was assessed using a chi-square test for categorical and an independent-samples t-test for continuous variables. We applied the Benjamini–Hochberg false discovery rate (FDR) correction across all subgroup–attribute pairs and considered results significant at $p < 0.05$. Only attributes with both significant results and an absolute effect size of at least 0.3 were retained for visualization.
All significant standardized effect sizes were compiled into a heatmap (Fig.~\ref{fig:heatmap}), where each column corresponds to a discovered subgroup and each row to a metadata attribute. Colors represent the direction and strength of association after standardization within metric type (Cohen’s h or d). Positive values indicate attributes that are more common or higher within a subgroup, and negative values indicate attributes that are less common or lower.

\subsection*{Synthetic scenario with generated artifacts} \label{sec:synth}

As a sanity check, we evaluate subgroup discovery in a simulated scenario where ground truth subgroups and subgroup performances are known. To simulate performance disparities, we add artifacts spuriously correlated with the positive disease label, similar to standard practice in shortcut learning research \cite{roschewitz2024automatic,bayasi2024biaspruner,bissoto2023even,sun2023right}. We adapt CheXpertPlus with two clinically-inspired artifacts following previous work~\cite{sun2023right} (Fig.\,\ref{fig:graphical-abstract}\,\textbf{b}): a \textit{hospital tag} on the bottom left, and vertical lines of \textit{hyperintense signal}.

The two artifacts are independently correlated with the label: each is inserted into positive samples with probability $p$ and into negative samples with probability $1-p$, yielding four ground-truth subgroups.
Training and validation sets are generated from this biased version of the data and used to train the target classification model and select its hyperparameters.
The validation set is also used to fit DOMINO.
For testing, we use an unbiased test set ($p=0.5$), facilitating fair comparisons across training bias levels.

\subsection*{Experimental setup}

For all experiments, we trained ResNet-50~\cite{he2016resnet} classification models using SGD and searched over learning rates of $\{10^{-5}, 10^{-4}, 10^{-3}\}$ with the weight decay of $10^{-4}$. Models were selected based on validation balanced accuracy. We chose the task of ``cardiomegaly vs. all''.
For subgroup discovery, we always used 15 subgroups. The external model is the pretrained CLIP~\cite{clip} for all scenarios, and we included BiomedCLIP~\cite{zhang2024biomedclip} for our real-world experiments. We used the publicly available versions of these pretrained models from the OpenAI repository and HuggingFace, respectively.
In the synthetic scenario, we considered hyperintensities as a known attribute and hospital tags as a hidden stratification and set the bias level as $p=0.8$.
%We use accuracy as our primary metric for measuring performance gaps and subgroup performances.

\subsection*{Perspective: regulatory implications}

Subgroup discovery has direct implications for regulatory oversight of clinical AI systems. To illustrate this connection, we consider how a chest X-ray classification tool developed as Software as a Medical Device would operate under the Medical Device Regulation (MDR). The MDR requires manufacturers to maintain a post-market surveillance system that continually assesses model performance and safety throughout the device lifetime (Article 83(1) MDR). This includes procedures for identifying situations in which the device may place patients at increased risk and for taking appropriate action when such concerns arise (Article 83(4) MDR). Subgroup discovery supports these obligations by revealing groups of patients for whom performance differs substantially from the overall model behaviour. It complements rather than replaces evaluations based on predefined socio-demographic groups, as it can uncover additional patient clusters with meaningful performance gaps, thereby strengthening trustworthiness.

As part of this surveillance process, the manufacturer will collect real-world data from routine clinical use. Subgroup discovery can be applied to model outputs on the validation dataset to partition samples into groups that show distinct performance profiles. This method requires only model outputs and a feature representation extracted by a separate model, so it can be deployed without retraining or access to model internals. Once subgroups are defined, the manufacturer can infer subgroup membership for new patients and determine whether an individual belongs to a group with lower validated accuracy. If this occurs, it can serve as an early warning signal that the model may not be reliable for that specific case. The associated metadata profiles provide additional context, enabling clinicians to assess whether the patient exhibits characteristics associated with poor performance and to decide whether to use or override the model output.

If an alert is raised because a new patient belongs to a subgroup with lower validated performance, it would trigger the procedures described in the manufacturer’s post-market surveillance plan (Article 84 of the MDR). The manufacturer would assess whether these alerts reflect isolated cases or indicate a broader limitation of the model. 
If corrective measures are required, such as model updates, our bias mitigation experiment shows that subgroup discovery can supply the labels needed for targeted retraining that improves performance for a clinically important group.

\subsection*{Data availability}
The CheXpertPlus dataset used in the current study is publicly available at \url{https://aimi.stanford.edu/datasets/chexpert-plus}. A list of CheXpert images labeled for the presence of pacemakers, which we used for validation of our method, is publicly available at \url{https://github.com/nina-weng/FastDiME_Med/blob/master/shortcut_annotations/che_pm_shortcut_labels.csv}.

\subsection*{Code availability}
The code for training and evaluating the models used in this work is available at \url{https://github.com/alceubissoto/hidden-sg-disparities}.

\subsubsection*{Acknowledgments} 

Calculations were performed on UBELIX (https://www.id.unibe.ch/hpc), the HPC cluster at the University of Bern.
This work was supported by the Diabetes Center Berne (AB, TH, TF, LK); the Carl Zeiss Foundation in the project ``Certification and Foundations of Safe Machine Learning Systems in Healthcare'' (SS); the Deutsche Forschungsgemeinschaft (DFG), EXC number 2064/1, Project number 39072764 (SS, CB); strategic funding of the medical faculty of the University of Bern (TH);  DTU Compute, the Technical University of Denmark (NW); Otto Mønsteds Fond (NW); IDA and Berg-Nielsen Study and Support Fund (NW); and DTU Travel Stipend (NW).

\subsubsection*{Author Contributions}
A.B. and L.K. designed the study. A.B. performed the experiments and analyses and wrote the first draft of the manuscript. L.K. supervised the project and acquired funding. T.H., T.F., and N.W. contributed to methodological discussions. All authors (A.B., T.H., T.F., N.W., S.S., C.B., L.K.) contributed to the interpretation of the results and reviewed, edited, and approved the final manuscript.

\subsubsection*{Competing Interests}
The authors declare no competing interests.

\bibliographystyle{elsarticle-num} \bibliography{references.bib}

@article{raoof2012interpretation,
  title={Interpretation of plain chest roentgenogram},
  author={Raoof, Suhail and Feigin, David and Sung, Arthur and Raoof, Sabiha and Irugulpati, Lavanya and Rosenow III, Edward C},
  journal={Chest},
  volume={141},
  number={2},
  pages={545--558},
  year={2012},
  publisher={Elsevier}
}

@misc{euaiact2025,
title = {Regulation (EU) 2024/1689 of the European Parliament and of the Council of 13 June 2024 laying down harmonised rules on artificial intelligence and amending Regulations (EC) No 300/2008, (EU) No 167/2013, (EU) No 168/2013, (EU) 2018/858, (EU) 2018/1139 and (EU) 2019/2144 and Directives 2014/90/EU, (EU) 2016/797 and (EU) 2020/1828 (Artificial Intelligence Act) (Text with EEA relevance)},
number = {1689},
year = {2024},
month = {Jul},
}

@article{nauta2023anecdotal,
  title={From anecdotal evidence to quantitative evaluation methods: A systematic review on evaluating explainable ai},
  author={Nauta, Meike and Trienes, Jan and Pathak, Shreyasi and Nguyen, Elisa and Peters, Michelle and Schmitt, Yasmin and Schl{\"o}tterer, J{\"o}rg and Van Keulen, Maurice and Seifert, Christin},
  journal={ACM Computing Surveys},
  volume={55},
  number={13s},
  pages={1--42},
  year={2023},
  publisher={ACM New York, NY}
}

@inproceedings{bissoto2025subgroup,
  title={Subgroup Performance Analysis in Hidden Stratifications},
  author={Bissoto, Alceu and Hoang, Trung-Dung and Fl{\"u}hmann, Tim and Sun, Susu and Baumgartner, Christian F and Koch, Lisa M},
  booktitle={International Conference on Medical Image Computing and Computer-Assisted Intervention (MICCAI)},
  pages={594--603},
  year={2025},
  organization={Springer}
}

@inproceedings{weng2024fast,
        title={Fast diffusion-based counterfactuals for shortcut removal and generation},
        author={Weng, Nina and Pegios, Paraskevas and Petersen, Eike and Feragen, Aasa and Bigdeli, Siavash},
        booktitle={European Conference on Computer Vision (ECCV)},
        pages={338--357},
        year={2024},
        organization={Springer}
      }

@inproceedings{
alloula2025subgroups,
title={Subgroups Matter for Robust Bias Mitigation},
author={Anissa Alloula and Charles Jones and Ben Glocker and Bartlomiej Papiez},
booktitle={International Conference on Machine Learning (ICML)},
year={2025},
url={https://openreview.net/forum?id=P0RkH1RT5z}
}

@article{zack2024assessing,
  title={Assessing the potential of GPT-4 to perpetuate racial and gender biases in health care: a model evaluation study},
  author={Zack, Travis and Lehman, Eric and Suzgun, Mirac and Rodriguez, Jorge A and Celi, Leo Anthony and Gichoya, Judy and Jurafsky, Dan and Szolovits, Peter and Bates, David W and Abdulnour, Raja-Elie E and others},
  journal={The Lancet Digital Health},
  volume={6},
  number={1},
  pages={e12--e22},
  year={2024},
  publisher={Elsevier}
}

@article{collins2024tripod+,
  title={TRIPOD+ AI statement: updated guidance for reporting clinical prediction models that use regression or machine learning methods},
  author={Collins, Gary S and Moons, Karel GM and Dhiman, Paula and Riley, Richard D and Beam, Andrew L and Van Calster, Ben and Ghassemi, Marzyeh and Liu, Xiaoxuan and Reitsma, Johannes B and Van Smeden, Maarten and others},
  journal={bmj},
  volume={385},
  year={2024},
  publisher={British Medical Journal Publishing Group}
}

@article{subbaswamy2024data,
  title={A data-driven framework for identifying patient subgroups on which an AI/machine learning model may underperform},
  author={Subbaswamy, Adarsh and Sahiner, Berkman and Petrick, Nicholas and Pai, Vinay and Adams, Roy and Diamond, Matthew C and Saria, Suchi},
  journal={npj Digital Medicine},
  volume={7},
  number={1},
  pages={334},
  year={2024},
  publisher={Nature Publishing Group UK London}
}

@article{lekadir2025future,
  title={FUTURE-AI: international consensus guideline for trustworthy and deployable artificial intelligence in healthcare},
  author={Lekadir, Karim and Frangi, Alejandro F and Porras, Antonio R and Glocker, Ben and Cintas, Celia and Langlotz, Curtis P and Weicken, Eva and Asselbergs, Folkert W and Prior, Fred and Collins, Gary S and others},
  journal={bmj},
  volume={388},
  year={2025},
  publisher={British Medical Journal Publishing Group}
}

@article{moons2025probast,
  title={PROBAST+ AI: an updated quality, risk of bias, and applicability assessment tool for prediction models using regression or artificial intelligence methods},
  author={Moons, Karel GM and Damen, Johanna AA and Kaul, Tabea and Hooft, Lotty and Navarro, Constanza Andaur and Dhiman, Paula and Beam, Andrew L and Van Calster, Ben and Celi, Leo Anthony and Denaxas, Spiros and others},
  journal={bmj},
  volume={388},
  year={2025},
  publisher={British Medical Journal Publishing Group}
}

@article{muralidharan2024scoping,
  title={A scoping review of reporting gaps in FDA-approved AI medical devices},
  author={Muralidharan, Vijaytha and Adewale, Boluwatife Adeleye and Huang, Caroline J and Nta, Mfon Thelma and Ademiju, Peter Oluwaduyilemi and Pathmarajah, Pirunthan and Hang, Man Kien and Adesanya, Oluwafolajimi and Abdullateef, Ridwanullah Olamide and Babatunde, Abdulhammed Opeyemi and others},
  journal={NPJ Digital Medicine},
  volume={7},
  number={1},
  pages={273},
  year={2024},
  publisher={Nature Publishing Group UK London}
}

@misc{fda2025,
  title={AI-Enabled Medical Devices},
  author={{U.S. Food and Drug Administration}},
  year={2025},
  howpublished={\url{https://www.fda.gov/medical-devices/software-medical-device-samd/artificial-intelligence-enabled-medical-devices}},
  note={Accessed: 28 August 2025}
}

@article{degrave2021ai,
  title={AI for radiographic COVID-19 detection selects shortcuts over signal},
  author={DeGrave, Alex J and Janizek, Joseph D and Lee, Su-In},
  journal={Nature Machine Intelligence},
  volume={3},
  number={7},
  pages={610--619},
  year={2021},
  publisher={Nature Publishing Group UK London}
}

@article{larrazabal2020gender,
  title={Gender imbalance in medical imaging datasets produces biased classifiers for computer-aided diagnosis},
  author={Larrazabal, Agostina J and Nieto, Nicol{\'a}s and Peterson, Victoria and Milone, Diego H and Ferrante, Enzo},
  journal={Proceedings of the National Academy of Sciences},
  volume={117},
  number={23},
  pages={12592--12594},
  year={2020},
  publisher={National Academy of Sciences}
}

@inproceedings{jimenez2023detecting,
  title={Detecting shortcuts in medical images-a case study in chest x-rays},
  author={Jim{\'e}nez-S{\'a}nchez, Amelia and Juodelyte, Dovile and Chamberlain, Bethany and Cheplygina, Veronika},
  booktitle={2023 IEEE 20th International Symposium on Biomedical Imaging (ISBI)},
  pages={1--5},
  year={2023},
  organization={IEEE}
}

@inproceedings{weng2023sex,
  title={Are sex-based physiological differences the cause of gender bias for chest x-ray diagnosis?},
  author={Weng, Nina and Bigdeli, Siavash and Petersen, Eike and Feragen, Aasa},
  booktitle={Workshop on Clinical Image-Based Procedures},
  pages={142--152},
  year={2023},
  organization={Springer}
}

@inproceedings{olesen2024slicing,
  title={Slicing through bias: explaining performance gaps in medical image analysis using slice discovery methods},
  author={Olesen, Vincent and Weng, Nina and Feragen, Aasa and Petersen, Eike},
  booktitle={MICCAI Workshop on Fairness of AI in Medical Imaging},
  pages={3--13},
  year={2024},
  organization={Springer}
}

@article{seyyed2021underdiagnosis,
  title={Underdiagnosis bias of artificial intelligence algorithms applied to chest radiographs in under-served patient populations},
  author={Seyyed-Kalantari, Laleh and Zhang, Haoran and McDermott, Matthew BA and Chen, Irene Y and Ghassemi, Marzyeh},
  journal={Nature medicine},
  volume={27},
  number={12},
  pages={2176--2182},
  year={2021},
  publisher={Nature Publishing Group US New York}
}

@article{zhang2024biomedclip,
  title={A Multimodal Biomedical Foundation Model Trained from Fifteen Million Image–Text Pairs},
  author={Sheng Zhang and Yanbo Xu and Naoto Usuyama and Hanwen Xu and Jaspreet Bagga and Robert Tinn and Sam Preston and Rajesh Rao and Mu Wei and Naveen Valluri and Cliff Wong and Andrea Tupini and Yu Wang and Matt Mazzola and Swadheen Shukla and Lars Liden and Jianfeng Gao and Angela Crabtree and Brian Piening and Carlo Bifulco and Matthew P. Lungren and Tristan Naumann and Sheng Wang and Hoifung Poon},
  journal={NEJM AI},
  year={2024},
  volume={2},
  number={1},
  doi={10.1056/AIoa2400640},
  url={https://ai.nejm.org/doi/full/10.1056/AIoa2400640}
}

@article{roschewitz2024automatic,
  title={Automatic dataset shift identification to support root cause analysis of AI performance drift},
  author={Roschewitz, M{\'e}lanie and Mehta, Raghav and Jones, Charles and Glocker, Ben},
  journal={arXiv preprint arXiv:2411.07940},
  year={2024}
}

@inproceedings{christodoulou2024confidence,
  title={Confidence intervals uncovered: Are we ready for real-world medical imaging AI?},
  author={Christodoulou, Evangelia and Reinke, Annika and Houhou, Rola and Kalinowski, Piotr and Erkan, Selen and Sudre, Carole H and Burgos, Ninon and Boutaj, Sofi{\`e}ne and Loizillon, Sophie and Solal, Ma{\"e}lys and others},
  booktitle={International Conference on Medical Image Computing and Computer-Assisted Intervention (MICCAI)},
  pages={124--132},
  year={2024},
  organization={Springer}
}

@inproceedings{bayasi2024biaspruner,
  title={Biaspruner: Debiased continual learning for medical image classification},
  author={Bayasi, Nourhan and Fayyad, Jamil and Bissoto, Alceu and Hamarneh, Ghassan and Garbi, Rafeef},
  booktitle={International Conference on Medical Image Computing and Computer-Assisted Intervention (MICCAI)},
  pages={90--101},
  year={2024},
  organization={Springer}
}

@inproceedings{sagawa2019distributionally,
  title={Distributionally robust neural networks for group shifts: On the importance of regularization for worst-case generalization},
  author={Sagawa, S. and Koh, P. W and Hashimoto, T. B and Liang, P.},
  booktitle={International Conference on Learning Representations (ICLR)},
  year={2020}
}

@inproceedings{irvin2019chexpert,
  title={Chexpert: A large chest radiograph dataset with uncertainty labels and expert comparison},
  author={Irvin, J. and Rajpurkar, P. and Ko, M. and Yu, Y. and Ciurea-Ilcus, S. and Chute, C. and Marklund, H. and Haghgoo, B. and Ball, R. and Shpanskaya, K. and others},
  booktitle={AAAI Conference on Artificial Intelligence},
  year={2019}
}

@article{bissoto2023even,
  title={Even Small Correlation and Diversity Shifts Pose Dataset-Bias Issues},
  author={Bissoto, A. and Barata, C. and Valle, E. and Avila, S.},
  journal={Pattern Recognition Letters},
  year={2024},
  volume={45},
}

@inproceedings{clip,
  title={Learning transferable visual models from natural language supervision},
  author={Radford, A. and Kim, J. W. and Hallacy, C. and Ramesh, A. and Goh, G. and Agarwal, S. and Sastry, G. and Askell, A. and Mishkin, P. and Clark, J. and others},
  booktitle={International Conference on Machine Learning (ICML)},
  pages={8748--8763},
  year={2021},
  organization={PMLR}
}

@article{chambon2024chexpert,
  title={CheXpert Plus: Augmenting a Large Chest X-ray Dataset with Text Radiology Reports, Patient Demographics and Additional Image Formats},
  author={Chambon, Pierre and Delbrouck, Jean-Benoit and Sounack, Thomas and Huang, Shih-Cheng and Chen, Zhihong and Varma, Maya and Truong, Steven QH and Chuong, Chu The and Langlotz, Curtis P},
  journal={arXiv preprint arXiv:2405.19538},
  year={2024}
}

@article{koch2024postmarket,
   author = {Lisa M Koch and Christian F Baumgartner and Philipp Berens},
   journal = {npj Digital Medicine},
   title = {Distribution Shift Detection for the Postmarket Surveillance of Medical AI Algorithms: A Retrospective Simulation Study},
   year = {2024},
}

@inproceedings{sun2023right,
   author = {Susu Sun and Lisa M. Koch and Christian F. Baumgartner},
   title = {Right for the Wrong Reason: Can Interpretable ML Techniques Detect Spurious Correlations?},
    booktitle = {Proc. Medical Image Computing and Computer Assisted Interventions MICCAI},
    year = {2023}
}

@inproceedings{
eyuboglu2022domino,
title={Domino: Discovering Systematic Errors with Cross-Modal Embeddings},
author={Sabri Eyuboglu and Maya Varma and Khaled Kamal Saab and Jean-Benoit Delbrouck and Christopher Lee-Messer and Jared Dunnmon and James Zou and Christopher Re},
booktitle={International Conference on Learning Representations},
year={2022},
}

@article{finlayson2021ClinicianDatasetShift,
	title = {The {Clinician} and {Dataset} {Shift} in {Artificial} {Intelligence}},
	volume = {385},
	number = {3},
	journal = {New England Journal of Medicine},
	author = {Finlayson, Samuel G. and Subbaswamy, Adarsh and Singh, Karandeep and Bowers, John and Kupke, Annabel and Zittrain, Jonathan and Kohane, Isaac S. and Saria, Suchi},
	month = jul,
	year = {2021},
	pages = {283--286},
}

@INPROCEEDINGS{he2016resnet,
  author={He, Kaiming and Zhang, Xiangyu and Ren, Shaoqing and Sun, Jian},
  booktitle={Proc. Conference on Computer Vision and Pattern Recognition (CVPR)}, 
  title={Deep Residual Learning for Image Recognition}, 
  year={2016},
}

@inproceedings{oakden2020hiddenStratification,
author = {Oakden-Rayner, L and Dunnmon, Jared and Carneiro, Gustavo and Re, Christopher},
title = {Hidden Stratification Causes Clinically Meaningful Failures in Machine Learning for Medical Imaging},
year = {2020},
booktitle = {ACM Conference on Health, Inference, and Learning},
}

%% else use the following coding to input the bibitems directly in the
%% TeX file.

% \begin{thebibliography}{00}

% %% \bibitem[Author(year)]{label}
% %% Text of bibliographic item

% \bibitem[ ()]{}

% \end{thebibliography}

%\section*{Figure Legends}

\listoffigures

\end{document}